\documentclass[conference]{IEEEtran}
\IEEEoverridecommandlockouts

\usepackage{amsmath,amssymb,amsfonts}
\usepackage{algorithmic}
\usepackage{graphicx}
\usepackage{textcomp}
\usepackage{xcolor}

\usepackage{CJKutf8}
\usepackage{CJK}

\usepackage{tabularx}
\usepackage{booktabs}
\usepackage{url}

\def\BibTeX{{\rm B\kern-.05em{\sc i\kern-.025em b}\kern-.08em
    T\kern-.1667em\lower.7ex\hbox{E}\kern-.125emX}}
\begin{document}

\title{Hard-Synth: Synthesizing Diverse Hard Samples \\for ASR using Zero-Shot TTS and LLM}

\author{Jiawei Yu$^{1*}$, Yuang Li$^{2*}$, Xiaosong Qiao$^{2}$, Huan Zhao$^{2}$, Xiaofeng Zhao$^{2}$,\\ Wei Tang$^{2}$, Min Zhang$^{2}$, Hao Yang$^{2}$, Jinsong Su$^{1}$\thanks{* denotes equal contribution to this work.}\\ $^{1}$School of Informatics, Xiamen University, China\\
$^{2}$Huawei Translation Services Center, China\\
\small{yujiawei@stu.xmu.edu.cn} liyuang3@huawei.com}

\maketitle

\begin{abstract}
Text-to-speech (TTS) models have been widely adopted to enhance automatic speech recognition (ASR) systems using text-only corpora, thereby reducing the cost of labeling real speech data. Existing research primarily utilizes additional text data and predefined speech styles supported by TTS models. In this paper, we propose Hard-Synth, a novel ASR data augmentation method that leverages large language models (LLMs) and advanced zero-shot TTS. Our approach employs LLMs to generate diverse in-domain text through rewriting, without relying on additional text data. Rather than using predefined speech styles, we introduce a hard prompt selection method with zero-shot TTS to clone speech styles that the ASR model finds challenging to recognize. Experiments demonstrate that Hard-Synth significantly enhances the Conformer model, achieving relative word error rate (WER) reductions of 6.5\%/4.4\% on LibriSpeech dev/test-other subsets. Additionally, we show that Hard-Synth is data-efficient and capable of reducing bias in ASR.

\end{abstract}
\section{Introduction}

The mutual evolution of automatic speech recognition (ASR)~\cite{chorowski2015attention, graves2006connectionist, graves2012sequence} and text-to-speech (TTS) technologies~\cite{ju2024naturalspeech, wang2023neural} has significantly advanced both fields. Modern TTS models can generate speech signals indistinguishable from real recordings~\cite{muller2022human, cooke2024good}, making them ideal for augmenting ASR training sets. Most research focuses on using text-only data and TTS models for domain adaptation~\cite{DBLP:conf/emnlp/SuFSCL24, DBLP:conf/interspeech/BataevKSLG23}, relying solely on target-domain text without real corresponding audio. Researchers have shown that synthetic audio can significantly improve the recall of out-of-vocabulary (OOV)~\cite{DBLP:conf/icassp/ZhengLGW21} and entity words~\cite{liang2023improving}. However, the synthetic-to-real gap can lead to performance degradation. To address this, mix-training, weight regularization such as elastic weight consolidation~\cite{DBLP:conf/icassp/ZhengLGW21}, and the use of partially different parameters for synthetic and real data are employed~\cite{DBLP:conf/emnlp/SuFSCL24, DBLP:conf/icassp/SYNT++}.

Zero-shot TTS can replicate a single utterance and produce speech signals with similar speaking speed, emotion, tone color, etc. Both autoregressive (AR) models, such as VALLE~\cite{wang2023neural} and VoiceCraft~\cite{peng2024voicecraft}, and non-autoregressive (NAR) models, such as NaturalSpeech3~\cite{ju2024naturalspeech} and F5-TTS~\cite{chen-etal-2024-f5tts}, have been proposed. Compared to conventional TTS models, zero-shot TTS enhances generation diversity and controllability. Consequently, zero-shot TTS can be employed to generate personalized speech for speaker adaptation~\cite{DBLP:conf/icassp/YangHCKT23, DBLP:conf/icassp/0028HWGLG20, DBLP:conf/interspeech/0028LHWGG20}. Additionally, it can be utilized to expand the training data for low-resource ASR tasks, such as those involving minority languages~\cite{yang2024enhancing} and dysarthric speech~\cite{DBLP:conf/icassp/SoleymanpourJSB22}. In this paper, we use zero-shot TTS to clone the speech style of hard utterances.

Large language models (LLMs)~\cite{gpt2023gpt, zeng2023glm130b, touvron2023llama} have shown remarkable performance across diverse natural language processing tasks, such as ASR error correction~\cite{ma2023generative}, text rewriting~\cite{Shu_2024}, and grammar correction~\cite{Fan_2023}. Recently, LLMs have gained traction for ASR data augmentation by creating text corpora for subsequent use in TTS models. For instance, LLMs can create Arabic-English code-switching data~\cite{alharbi2024leveraging} and communication data~\cite{DBLP:journals/corr/abs-2408-09215}, as well as text with specific entity words as instructed~\cite{DBLP:conf/icassp/SuHKVCYMT24}. In this study, we adopt a simple yet effective approach to instruct the LLM to rewrite the sentences in the original training set.

In this paper, we propose Hard-Synth which leverages LLMs and zero-shot TTS to synthesize diverse hard samples to augment the ASR training set. Our approach utilizes LLMs to rewrite the original text in the training set, generating text with the same meaning but with different wording and structure. Additionally, we introduce a hard prompt selection method that employs a weak ASR model to identify difficult utterances, which are later used as audio prompts for zero-shot TTS. LLM rewriting ensures that the generated text data remains within the same domain as the training set. Cloning hard utterances using zero-shot TTS can balance the training set since the acoustic properties of these utterances are normally less frequent. Experiments on the LibriSpeech~\cite{panayotov2015librispeech} dataset demonstrate that Hard-Synth consistently enhances the performance of Transformer~\cite{vaswani2017attention} and Conformer~\cite{gulati2020conformer} models. For instance, Hard-Synth achieves relative word error rate (WER) reductions of 6.5\% and 4.4\% on the dev/test-other subsets for the Conformer model. Furthermore, Hard-Synth is data-efficient, achieving these improvements with only 16.15 hours of synthetic data, which is just 16\% of the real data. This efficiency indicates that Hard-Synth is cost-effective in data creation and imposes negligible additional computational demands during training. Additionally, comprehensive analyses reveal that AR TTS models are more suitable for our framework compared to NAR TTS models, and Hard-Synth effectively mitigates biases in ASR, such as gender bias and performance variations between speakers.
\section{Methodology}

\subsection{System Overview}

\begin{figure}[t]
  \centering
  \includegraphics[width=\linewidth]{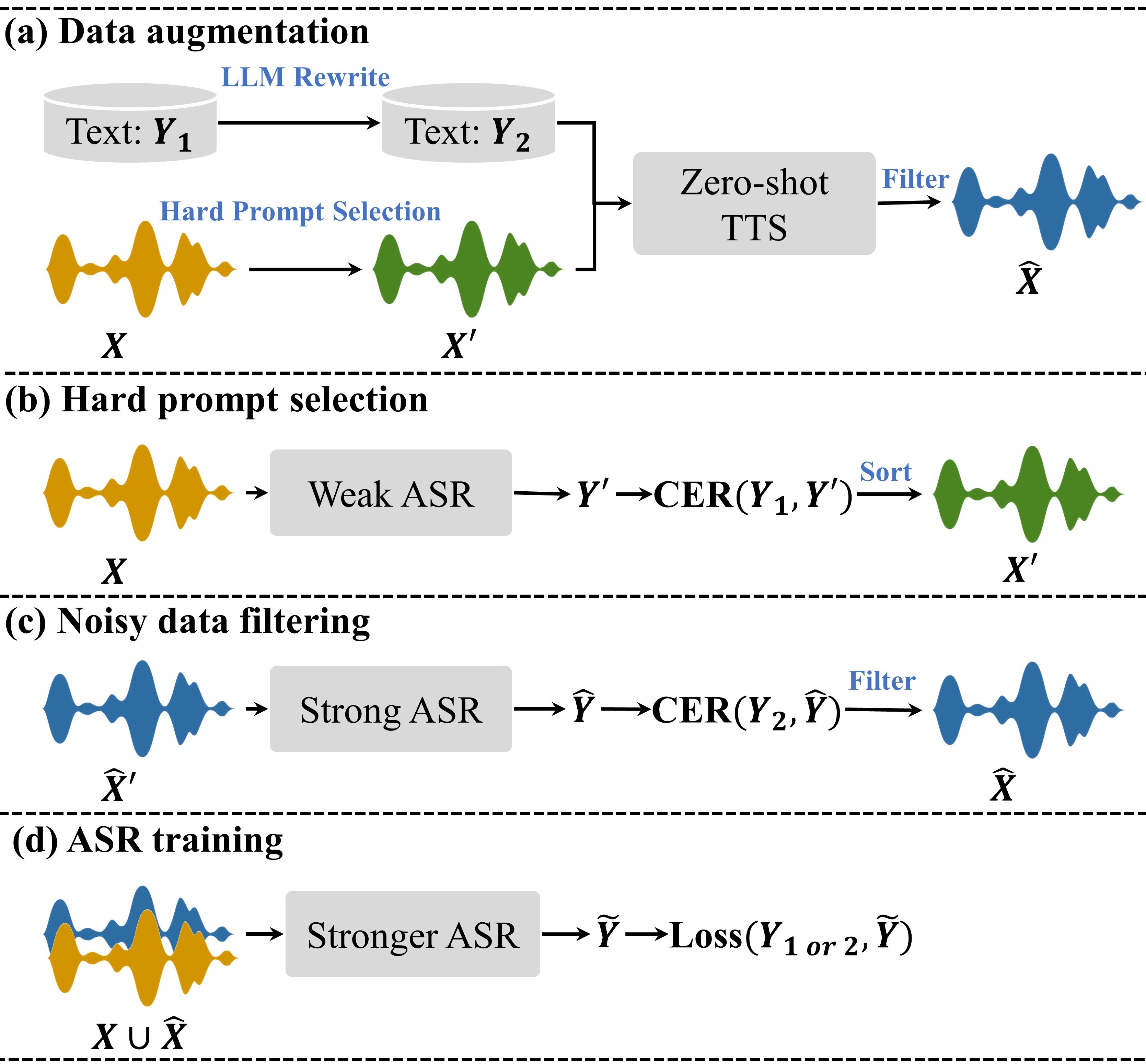}
  \caption{System diagram of Hard-Synth. (a) Zero-shot TTS is employed with hard audio prompts ($\mathbf{X}'$) and text generated by LLM ($\mathbf{Y}_2$) to produce synthetic speech signals ($\mathbf{\hat{X}}$). (b) Hard audio prompts are defined as speech samples that a weak ASR model finds difficult to recognize. (c) Before training, the synthetic speech signals are filtered based on CER. (d) Real and synthetic audio samples are then combined for the final training phase.}
  \label{fig:method}
\end{figure}

Figure~\ref{fig:method} provides an overview of the Hard-Synth methodology. As illustrated in Figure~\ref{fig:method} (a), the data augmentation pipeline comprises the generation of the text corpus $\mathbf{Y}_2$, the selection of hard speech prompts $\mathbf{X}'$, the synthesis and filtering of speech $\hat{\mathbf{X}}$. Specifically, $\mathbf{Y}_2$ is the LLM-rewritten version of the original training transcriptions $\mathbf{Y}_1$. $\mathbf{X}'$ is a subset of training speech signals $\mathbf{X}$, selected based on the character error rate (CER) of a weak ASR model. The zero-shot TTS model is utilized to synthesize speech signals containing the content of $\mathbf{Y}_2$ in the style of the hard audio prompts $\mathbf{X}'$. In other words, the semantic information of $\hat{\mathbf{X}}$ is derived from $\mathbf{Y}_2$, while the acoustic characteristics (e.g., accent, timbre, and environmental condition) are sourced from $\mathbf{X}'$. Finally, the combined dataset of real and synthetic speech signals is used to enhance the ASR model.

\subsection{Hard Prompt Selection}
\label{sec:hard}

 We define hard prompts as utterances that are particularly challenging for an ASR model to recognize. For this purpose, we prefer to use a weak ASR model that has not yet converged on the entire training set. The rationale is that an ASR model tends to overfit easy and high-resource samples quickly while underfitting difficult and low-resource samples in the early stages of training~\footnote{High-resource and low-resource indicate samples with common and rare acoustic characteristics respectively.}. If an ASR model has overfitted the training set, it would be infeasible to use it to identify hard samples.

Figure \ref{fig:method} (b) illustrates our hard prompt selection pipeline. We employ a weak ASR model trained for a limited number of epochs to transcribe the speech signals in the training set. To improve decoding efficiency and select hard utterances, we utilize connectionist temporal classification (CTC) with greedy decoding, as opposed to the more accurate yet computationally expensive attention-based decoding with beam search. Then, we compute the CER between the hypotheses $\mathbf{Y}'$ and references $\mathbf{Y}_1$. A higher CER indicates a greater level of difficulty for the ASR model and potentially identifies long-tail samples. For example, samples with high CERs can contain uncommon prosody and background noises. We choose CER over WER due to its finer granularity, which minimizes the influence of semantic content while emphasizing acoustic properties. Compared to CER, WER is more susceptible to the influence of text distribution and vocabulary, as a single incorrect character can result in a misrecognized word. Thus, WER is inferior for selecting hard audio prompts.

\subsection{Noisy data filtering}

Although the zero-shot TTS models are powerful, they can be occasionally unstable during decoding, especially for AR models. For instance, synthetic speech signals can contain long silences, missing words, and repeating words. This instability is akin to the phenomenon of hallucinations observed in LLMs. Consequently, to enhance the quality of synthetic data, we use an ASR model to transcribe the generated speech, ensuring that the speech content aligns with the input text (Figure~\ref{fig:method} (b)). In this context, we prefer a strong ASR model rather than a weak one. Hence, we select the ASR model that has converged on the original training set. For the same reason as outlined in Section~\ref{sec:hard}, we use the CER rather than the WER to assess speech quality. We filter out samples with a CER exceeding a predefined threshold.





\subsection{LLM Rewriting}

In Hard-Synth, our objective extends beyond only extending and balancing the training data by using hard audio prompts, which can be seen as data augmentation from an acoustic perspective. We also aim to enhance the diversity of semantic content. LLMs can be used to rewrite text effectively due to their semantic understanding and coherent generation capacities. Therefore, we employ LLMs to rewrite the text in the training set to improve the variety in sentence structures and vocabulary. The rewriting process can be regarded as a knowledge extraction procedure where the LLM serves as a knowledge base. By prompting the LLM with the original text, we instruct it to produce sentences using domain-relevant knowledge. By training the ASR system with such data, we can effectively transfer the LLM's knowledge to the internal language model of the ASR system, thereby enhancing its performance.

Table~\ref{table:prompt} shows the prompt for LLM rewriting. We instruct the LLM to act as a text rewriter and rewrite the given sentence without altering its meaning. The LLM is directed to provide the rewritten sentence directly, ensuring that the generated sentences maintain a consistent format. We employ a zero-shot prompt without providing paired examples as the LLM has already demonstrated strong performance. Some examples are provided in Table~\ref{table:example}. The rewritten sentences can be categorized into two types: paraphrasing (such as in example (a), where "hesitated a moment" is replaced with "paused briefly") and restructuring (as seen in example (b), where "before going to the sea" is repositioned to the end of the sentence). In practice, most sentences are a mixture of both paraphrasing and restructuring, as illustrated by the final example.





\begin{table}[t]
\caption{The Prompt for LLM rewriting.}
\centering
    \begin{tabular}{p{7cm}}
    \toprule
    You are a professional text rewriter.\\
    Please rewrite the following sentence without changing its meaning. Please give the rewritten sentence directly. \\
    Sentence: \{sentence\}\\
    \bottomrule
    \end{tabular}
    \label{table:prompt}
\end{table}

\begin{table}[t]
\caption{Examples of LLM rewriting. "1." represents the original sentence, and "2." indicates the generated sentence.}
\centering
    \begin{tabular}{p{1.9cm} | p{5.8cm}}
    \toprule
    \textbf{a}. Paraphrasing & 1. the girl hesitated a moment\\
    &2. She paused briefly\\
    \midrule
    \textbf{b}. Restructuring & 1. before going to sea said philip half smiling\\
    &2. Philip said half-smiling before going to sea\\
    \midrule
    \textbf{a} + \textbf{b} & 1. it's a glorious mission but also a dangerous one\\
    &2. This mission is both glorious and perilous\\
    \bottomrule
    \end{tabular}
    \label{table:example}
\end{table}

\section{Experimental Setups}

\subsection{Zero-shot TTS Models}

For Hard-Synth, we primarily use VoiceCraft~\cite{peng2024voicecraft}. In Section~\ref{sec:compare}, we compare it with F5-TTS~\cite{chen-etal-2024-f5tts} and demonstrate that, although F5-TTS generates higher quality audio, VoiceCraft excels in replicating the acoustic environment and speaking speed of the audio prompt. The brief introductions of VoiceCraft and F5-TTS are as follows:

\begin{itemize}
    \item \textbf{VoiceCraft}~\cite{peng2024voicecraft}: VoiceCraft utilizes sequence infilling for speech editing and zero-shot TTS by rearranging neural codec output tokens through a left-to-right language modeling approach. This process involves causal masking for AR continuation and infilling with bidirectional context, and delayed stacking for efficient multi-codebook modeling. VoiceCraft employs decoder-only Transformers and is trained using AR sequence prediction.
    \item \textbf{F5-TTS}~\cite{chen-etal-2024-f5tts}: F5-TTS is a fully NAR TTS system based on flow matching with the Diffusion Transformer. Unlike conventional TTS systems, it eliminates the need for intricate components such as duration models, text encoders, and phoneme alignment. Instead, text inputs are padded with filler tokens to align with the length of the input speech, and the denoising process is then applied to generate the speech.
\end{itemize}

\subsection{Synthetic Data Generation}

\begin{figure}[t]
  \centering
  \includegraphics[width=0.9\linewidth]{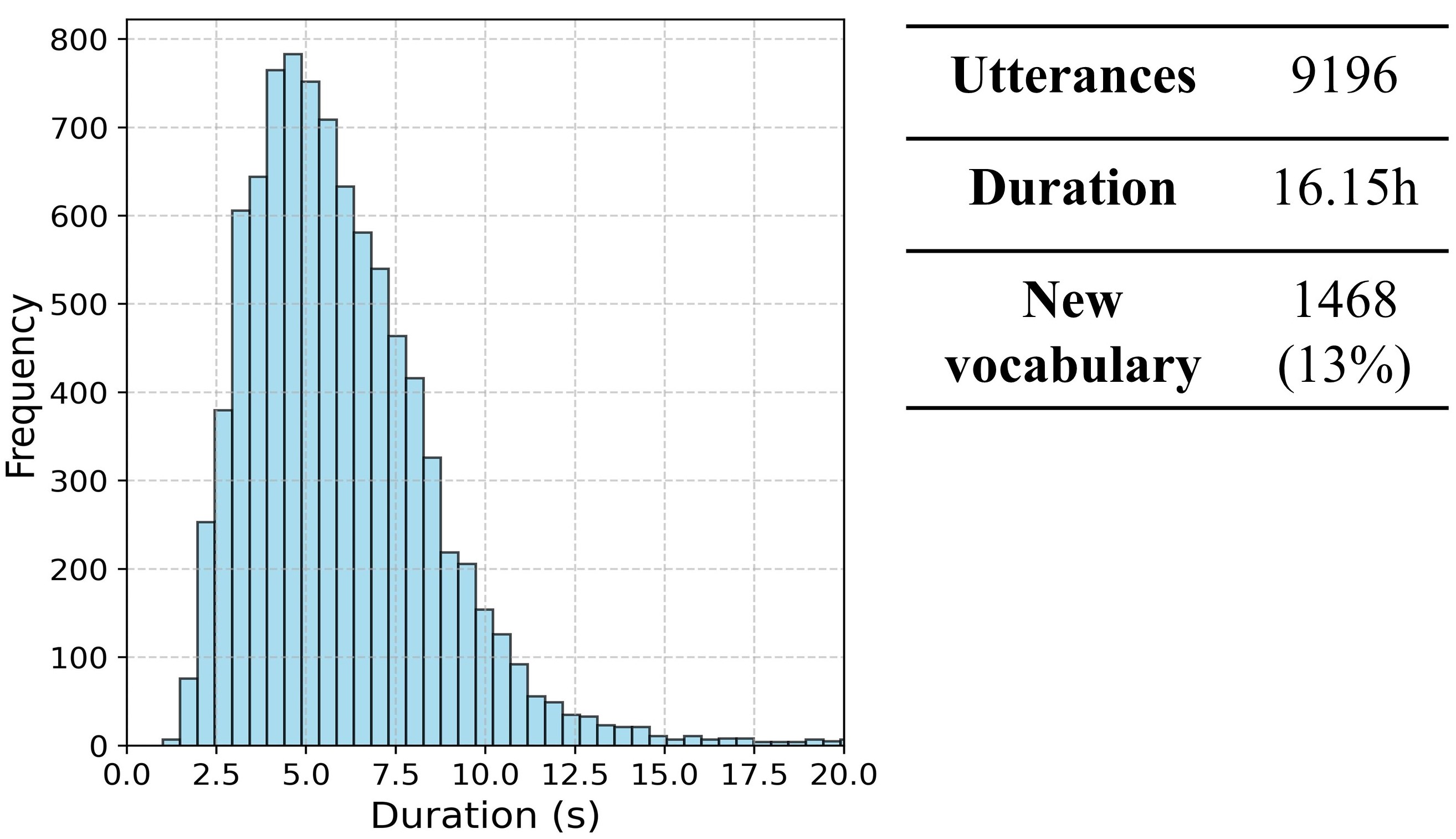}
  \caption{Statistics of the synthetic dataset.}
  \label{fig:data}
\end{figure}

We utilize the LibriSpeech-clean-100~\cite{panayotov2015librispeech} subset, consisting of 100 hours of speech data, for ASR training and data synthesis. The full dataset of 960 hours is not used, as we aim to simulate a low-resource scenario where data augmentation is more crucial. We utilize LLaMA3.1-8B~\cite{touvron2023llama} for text rewriting and employ an ASR model, trained for 15 epochs on the training set, as the weak ASR model for hard prompt selection. We sort the samples in descending order based on their CERs and select the top utterances with a total duration of 20 hours. Only audio clips longer than 3 seconds are retained as prompts, as shorter clips lack sufficient speech information, making it challenging for the TTS model to accurately reproduce the audio features of the original speech prompt. For noisy data filtering, we set the CER threshold to 10\%. As depicted in Figure~\ref{fig:data}, the synthetic dataset comprises 9,196 utterances with a total duration of 16.15 hours, introducing 13\% new vocabulary. The majority of these utterances are short, averaging around 6 seconds in length. For the ablation study, we also generate alternative variants of synthetic datasets, which will be discussed in Section~\ref{sec:ablation}.

\subsection{Model Training}

We follow the receipts in Espnet~\cite{watanabe2018espnet} to train the Transformer~\cite{vaswani2017attention} and Conformer~\cite{gulati2020conformer} models which contain 29.38 million and 34.23 million parameters, respectively. The models are trained using CTC and attention-based loss over 70 epochs, with a learning rate of 2e-3 and a warmup of 15,000 steps. The input features are Fbanks with a window length of 400 and a hop length of 160. SpecAug~\cite{park2019specaugment} is adopted for better generalizations. Post-training, the 10 best checkpoints are averaged, and joint decoding using the hybrid CTC/attention~\cite{watanabe2017hybrid} approach is performed with a beam size of 10.

\subsection{Evaluation}

We use WER to evaluate ASR performance. To assess audio quality and its similarity to the audio prompt, we employ: 1) MOSnet score~\cite{mosnet}; 2) Sim-spk: the cosine similarity between the speaker embeddings of the generated and prompt audio derived from WavLM-SV~\cite{chen2021wavlm}; 3) $\Delta$speed: the difference in speaking speed between the generated and prompt audio. We also identify the bias in ASR by computing: 1) $\Delta$WER-Gender: the difference in WER between male and female speakers; 2) Var-WER-Spk: the variance in WER across different speakers.
\begin{table}[t]
\caption{Performance of ASR models with Hard-Synth. '+ Synth (hard)' refers to the combination of real and synthetic speech data, with synthetic data generated using hard prompts and text randomly sampled from the training set. '+ Synth (hard + rewrite)' indicates that the LLM generates the synthetic transcriptions.}
\centering
\begin{tabular}{c | c c | c c | c} 
\toprule
& \multicolumn{2}{c|}{Dev} & \multicolumn{2}{c|}{Test} & \\
Training Data & clean & other & clean & other & Avg. \\
\midrule
\multicolumn{6}{c}{Transformer}\\
\midrule
Real & 7.92 & 19.70 & 8.39 & 20.13 & 14.04\\
+ Synth (hard) & 8.05 & 19.54 & 8.31 & 20.03 & 13.98\\
+ Synth (hard \& rewrite) & \textbf{7.67} & \textbf{19.30} & \textbf{7.92} & \textbf{19.30} & \textbf{13.55}\\
\midrule
\multicolumn{6}{c}{Conformer}\\
\midrule
Real & 6.46 & 17.47 & 6.54 & 17.43 & 11.98\\
+ Synth (hard) & \textbf{6.24} & 16.91 & 6.55 & 16.74 & 11.61\\
+ Synth (hard \& rewrite) & 6.32 & \textbf{16.34} & \textbf{6.39} & \textbf{16.67} & \textbf{11.43}\\
\bottomrule
\end{tabular}
\label{table:main}
\end{table}

\begin{table}[t]
\caption{Comparisons of synthetic data durations ($T$), filtering CER thresholds ($\gamma$), and hard/random audio prompts using the Conformer model with the LLM rewrite deactivated.}
\centering
\begin{tabular}{c | c c | c c | c} 
\toprule
& \multicolumn{2}{c|}{Dev} & \multicolumn{2}{c|}{Test} & \\
Parameters & clean & other & clean & other & Avg. \\
\midrule
\multicolumn{6}{c}{$\gamma=10\%$}\\
\midrule
$T$=8.62h & 6.38 & 17.11 & \textbf{6.51} & 17.16 & 11.79\\
$T$=15.38h & \textbf{6.24} & \textbf{16.91} & 6.55 & \textbf{16.74} & \textbf{11.61}\\
$T$=30.87h & 6.75 & 17.20 & 7.00 & 17.21 & 12.04\\
\midrule
\multicolumn{6}{c}{$T$=15.38h}\\
\midrule
$\gamma=5\%$ & 6.27 & \textbf{16.78} & \textbf{6.46} & 17.01 & 11.63\\
$\gamma=10\%$ & \textbf{6.24} & 16.91 & 6.55 & \textbf{16.74} & \textbf{11.61}\\
\midrule
\multicolumn{6}{c}{$\gamma=10\%$, $T$=15.38h}\\
\midrule
random prompt & 6.39 & \textbf{16.89} & 6.58 & 17.02 & 11.72\\
hard prompt & \textbf{6.24} & 16.91 & \textbf{6.55} & \textbf{16.74} & \textbf{11.61}\\
\bottomrule
\end{tabular}
\label{table:ablation}
\end{table}

\section{Experimental Results}

\subsection{Main Results}

As illustrated in Table~\ref{table:main}, Hard-Synth consistently enhances the performance of both the Transformer and Conformer models across all LibriSpeech subsets. The improvements are particularly notable in the "other" subsets compared to the "clean" subsets. For the Conformer model, the WER decreases from 17.47\% and 17.43\% to 16.34\% and 16.67\% on the dev-other and test-other subsets respectively. This is expected, as the "other" subsets contain more challenging samples than the "clean" subsets. Additionally, without LLM rewriting, using the selected hard audio prompts with randomly sampled text from the training set can still enhance ASR performance. For example, reducing the WER from 17.43\% to 16.74\% on the test-other subset for the Conformer model. This confirms the effectiveness of hard prompt selection since the augmentation focuses solely on acoustic aspects.

\begin{table}[t]
\caption{The performance comparison between VoiceCraft and F5-TTS with hard prompt selection and LLM rewriting.}
\centering
\begin{tabular}{c | c c | c c | c} 
\toprule
& \multicolumn{2}{c|}{Dev} & \multicolumn{2}{c|}{Test} & \\
TTS Model & clean & other & clean & other & Avg. \\
\midrule
F5-TTS & \textbf{6.01} & 16.87 & \textbf{6.32} & 17.01 & 11.55\\
VoiceCraft & 6.32 & \textbf{16.34} & 6.39 & \textbf{16.67} & \textbf{11.43}\\
\bottomrule
\end{tabular}
\label{table:tts1}
\end{table}

\begin{table}[t]
\caption{Comparisons between the synthetic speech of VoiceCraft and F5-TTS. The speech quality is measured by WER and MOSnet scores. The similarity with the prompt is measured by the similarity between speaker embeddings (Sim-spk) and the difference between speaking speeds ($\Delta$speed).}
\centering
\begin{tabular}{c | c c c c} 
\toprule
TTS Model & WER (\%) & MOSnet & Sim-spk & $\Delta$speed (words/s)\\
\midrule
F5-TTS & 3.57 & 3.46 & 0.960 & 1.46 \\
VoiceCraft & 4.59 & 3.26 & 0.949 & 1.35 \\
\bottomrule
\end{tabular}
\label{table:tts2}
\end{table}

\subsection{Ablation Study}
\label{sec:ablation}

Table~\ref{table:ablation} provides comparisons of synthetic speech durations, filtering CER thresholds, and hard/random audio prompts. The results indicate that a duration of 15.38 hours yields the best performance, surpassing the durations of 8.62 hours and 30.87 hours. This aligns with expectations, as a smaller amount of data may lack diversity, while a larger amount can degrade performance due to the synthetic-to-real gap. Additionally, the use of hard prompts is likely to alter the data distribution too much when increasing the duration of synthetic data. This observation further demonstrates the efficiency of Hard-Synth, where a modest quantity of synthetic data can yield significant improvements, thereby saving the resources required for data generation and ASR model training. When comparing CER thresholds of 5\% and 10\%, there is no noticeable difference. In the comparison between random and hard audio prompts, the hard prompts demonstrate superior performance, achieving a WER of 16.74\% on the test-other subset compared to 17.02\% with random prompts, proving the effectiveness of hard prompt selection.

\subsection{TTS Model Selection}
\label{sec:compare}

\begin{figure}[t]
  \centering
  \includegraphics[width=\linewidth]{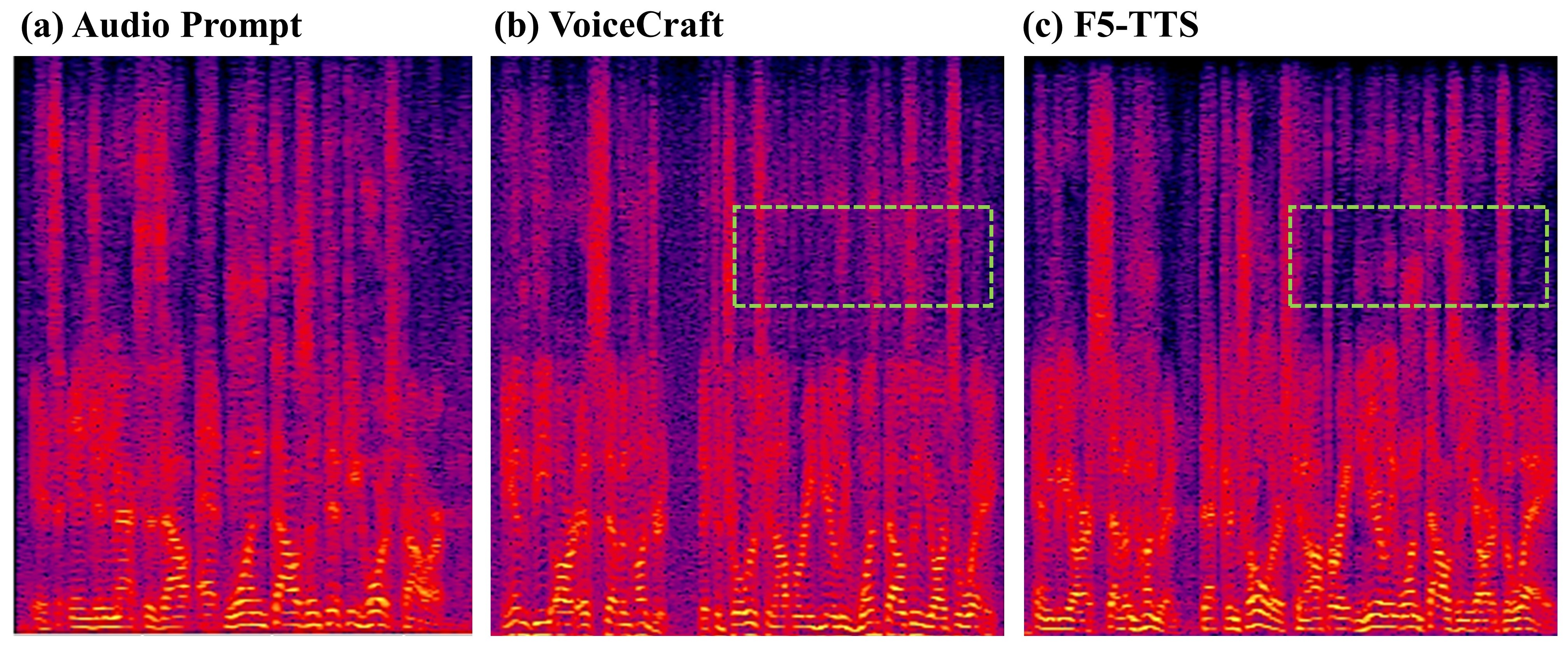}
  \caption{Visualizations of spectrograms. VoiceCraft excels in replicating the noise characteristics of the audio prompt compared to F5-TTS.}
  \label{fig:spec}
\end{figure}

F5-TTS~\cite{chen-etal-2024-f5tts} is the latest state-of-the-art zero-shot TTS model. However, as illustrated in Table~\ref{table:tts1}, when employed in Hard-Synth, its performance is slightly inferior to that of VoiceCraft on average (11.55\% compared to 11.34\%). Experiments reveal that F5-TTS achieves a lower WER on the "clean" subsets but a higher WER on the "other" subsets. This discrepancy arises because TTS models are typically designed to generate perceptually pleasing and clear speech signals, rather than challenging samples for ASR. VoiceCraft's AR generation better emulates the speech properties of the prompt. Conversely, the NAR inference and denoising steps of F5-TTS's diffusion process result in high-quality speech signals which are not the desired hard samples for Hard-Synth. As indicated in Table~\ref{table:tts2}, F5-TTS has better speech quality, evidenced by a lower WER, higher MOSnet score, and greater speaker similarity with the prompt. However, its speaking speed exhibits a greater deviation from the prompt compared to VoiceCraft. Figure~\ref{fig:spec} shows the spectrograms of cloned speech signals from VoiceCraft and F5-TTS when a noisy speech signal is used as the prompt. It is observed that VoiceCraft replicates noises similar to those in the prompt, whereas the signal generated by F5-TTS is relatively clean, as indicated by the dark areas in the spectrogram.

\begin{table}[t]
\caption{Evaluations of Bias in ASR. "$\Delta$WER-Gender" denotes the difference in WER between male and female speakers. "Var-WER-Spk" represents the variance in WER across different speakers.}
\centering
\begin{tabular}{c | c c | c c} 
\toprule
& \multicolumn{2}{c|}{$\Delta$WER-Gender} & \multicolumn{2}{c}{Var-WER-Spk} \\
& dev-other & test-other & dev-other & test-other\\
\midrule
Conformer & 3.07 & 2.06 & 116.1 & 86.5 \\
Hard-Synth & \textbf{2.86} & \textbf{1.66} & \textbf{99.4} & \textbf{79.5} \\
\bottomrule
\end{tabular}
\label{table:bias}
\end{table}

\subsection{Bias Reduction in ASR}

Bias in ASR systems exists in several areas, including gender, age, speech impairment, race, and accents~\cite{feng2021quantifying}. Hard-Synth offers an automated approach to mitigate these biases, as the hard samples typically originate from minority groups. Our observations indicate that the ASR model trained on LibriSpeech exhibits a higher WER for male speakers compared to female speakers. As shown in Table~\ref{table:bias}, Hard-Synth can reduce this bias, decreasing the WER gap from 2.06\% to 1.66\% on the test-other subset. We further examine the variance of WER among speakers and show that Hard-Synth consistently reduces such variance.

\section{Discussion}

\subsection{Relationship to SpecAug}

SpecAug~\cite{park2019specaugment} is the most widely adopted data augmentation method in ASR, which involves random masking of time frames and frequency bins of the Fbanks. This approach only modifies the speech features, similar to our method without LLM rewriting. Additionally, the random masking in SpecAug increases the difficulty of recognition, thereby mitigating the risk of overfitting. Similarly, our method also increases the recognition difficulty but in a more strategic manner by using a zero-shot TTS model. In all our experiments, we employ SpecAug and Hard-Synth concurrently, demonstrating that our method provides cumulative benefits when applied in conjunction with SpecAug.

\subsection{Preferred TTS Model}

The primary objective of TTS is to produce high-quality speech signals with good intelligibility. Consequently, the more advanced a TTS model is, the less suitable it may be for generating hard samples. An ideal zero-shot TTS model should generate "low-quality" speech signals if the audio prompt is of low quality. It should replicate all speech properties of the prompt, including prosody, timbre, noise, reverberation, accent, and other characteristics. Therefore, when developing zero-shot TTS models for ASR or emerging speech language models, researchers should focus not only on the accuracy of generation and speaker similarity with the prompt but also on a multi-perspective evaluation. These directions will help minimize the synthetic-to-real gap and allow for an increased proportion of synthetic data without the need for complex regularization strategies.

\subsection{Adaptation of Speech Foundation Models}

To eliminate the influence of pre-training data, we train the ASR model from scratch instead of using speech foundation models like Whisper~\cite{radford2022robust}. It is important to emphasize that our method is also applicable to pre-trained ASR models, allowing them to adapt to domains where they struggle to generalize. These domains typically involve speech with strong accents, high levels of background noise, and overlapping speakers. In such cases, the weak ASR model can be the pre-trained model itself, facilitating the selection of hard samples. However, these challenging samples remain difficult for TTS models to clone, which we identify as a direction for future research.
\section{Conclusion}

In conclusion, we propose a novel ASR data augmentation method, Hard-Synth, which leverages advanced zero-shot TTS models to clone hard audio prompts and employs LLM to enhance text diversity. Through comprehensive experiments, we demonstrate the superiority of our approach in terms of ASR performance, data efficiency, and bias reduction. Additionally, we provide a detailed analysis of hyperparameter selection and illustrate the advantages of using the AR TTS model over the NAR model. For future works, we will investigate more low-resource ASR tasks such as dysarthric ASR, and fine-tune the TTS model on hard audio samples.

\clearpage

\bibliographystyle{IEEEbib}
\bibliography{refs}

\begin{thebibliography}{10}

\bibitem{chorowski2015attention}
Jan Chorowski, Dzmitry Bahdanau, Dmitriy Serdyuk, Kyunghyun Cho, and Yoshua Bengio,
\newblock ``Attention-based models for speech recognition,''
\newblock in {\em Proc. of NeurIPS}, {Dec.} 2015, pp. 577--585.

\bibitem{graves2006connectionist}
Alex Graves, Santiago Fern{\'a}ndez, Faustino Gomez, and J{\"u}rgen Schmidhuber,
\newblock ``Connectionist temporal classification: labelling unsegmented sequence data with recurrent neural networks,''
\newblock in {\em Proc. of ICML}, {Jun.} 2006, pp. 369--376.

\bibitem{graves2012sequence}
Alex Graves,
\newblock ``Sequence transduction with recurrent neural networks,''
\newblock in {\em Proc. of ICML}, {Edinburgh, Scotland}, {Jun.} 2012.

\bibitem{ju2024naturalspeech}
Zeqian Ju, Yuancheng Wang, Kai Shen, Xu~Tan, Detai Xin, Dongchao Yang, Yanqing Liu, Yichong Leng, Kaitao Song, Siliang Tang, et~al.,
\newblock ``Naturalspeech 3: Zero-shot speech synthesis with factorized codec and diffusion models,''
\newblock {\em arXiv preprint arXiv:2403.03100}, 2024.

\bibitem{wang2023neural}
Chengyi Wang, Sanyuan Chen, Yu~Wu, Ziqiang Zhang, Long Zhou, Shujie Liu, Zhuo Chen, Yanqing Liu, Huaming Wang, Jinyu Li, et~al.,
\newblock ``Neural codec language models are zero-shot text to speech synthesizers,''
\newblock {\em arXiv preprint arXiv:2301.02111}, 2023.

\bibitem{muller2022human}
Nicolas~M M{\"u}ller, Karla Pizzi, and Jennifer Williams,
\newblock ``Human perception of audio deepfakes,''
\newblock in {\em Proceedings of the 1st International Workshop on Deepfake Detection for Audio Multimedia}, 2022, pp. 85--91.

\bibitem{cooke2024good}
Di~Cooke, Abigail Edwards, Sophia Barkoff, and Kathryn Kelly,
\newblock ``As good as a coin toss human detection of ai-generated images, videos, audio, and audiovisual stimuli,''
\newblock {\em arXiv preprint arXiv:2403.16760}, 2024.

\bibitem{DBLP:conf/emnlp/SuFSCL24}
Hsuan Su, Hua Farn, Fan{-}Yun Sun, Shang{-}Tse Chen, and Hung{-}yi Lee,
\newblock ``Task arithmetic can mitigate synthetic-to-real gap in automatic speech recognition,''
\newblock in {\em Proc. of EMNLP}, Yaser Al{-}Onaizan, Mohit Bansal, and Yun{-}Nung Chen, Eds., 2024.

\bibitem{DBLP:conf/interspeech/BataevKSLG23}
Vladimir Bataev, Roman Korostik, Evgeny Shabalin, Vitaly Lavrukhin, and Boris Ginsburg,
\newblock ``Text-only domain adaptation for end-to-end {ASR} using integrated text-to-mel-spectrogram generator,''
\newblock in {\em Proc. of Interspeech}, Naomi Harte, Julie Carson{-}Berndsen, and Gareth Jones, Eds., 2023.

\bibitem{DBLP:conf/icassp/ZhengLGW21}
Xianrui Zheng, Yulan Liu, Deniz Gunceler, and Daniel Willett,
\newblock ``Using synthetic audio to improve the recognition of out-of-vocabulary words in end-to-end asr systems,''
\newblock in {\em Proc. of ICASSP}, 2021.

\bibitem{liang2023improving}
Zheng Liang, Zheshu Song, Ziyang Ma, Chenpeng Du, Kai Yu, and Xie Chen,
\newblock ``Improving code-switching and named entity recognition in asr with speech editing based data augmentation,''
\newblock {\em arXiv preprint arXiv:2306.08588}, 2023.

\bibitem{DBLP:conf/icassp/SYNT++}
Ting{-}Yao Hu, Mohammadreza Armandpour, Ashish Shrivastava, Jen{-}Hao~Rick Chang, Hema Koppula, and Oncel Tuzel,
\newblock ``{SYNT++:} utilizing imperfect synthetic data to improve speech recognition,''
\newblock in {\em Proc. of ICASSP}, 2022.

\bibitem{peng2024voicecraft}
Puyuan Peng, Po-Yao Huang, Abdelrahman Mohamed, and David Harwath,
\newblock ``Voicecraft: Zero-shot speech editing and text-to-speech in the wild,''
\newblock {\em arXiv}, 2024.

\bibitem{chen-etal-2024-f5tts}
Yushen Chen, Zhikang Niu, Ziyang Ma, Keqi Deng, Chunhui Wang, Jian Zhao, Kai Yu, and Xie Chen,
\newblock ``F5-tts: A fairytaler that fakes fluent and faithful speech with flow matching,''
\newblock {\em arXiv preprint arXiv:2410.06885}, 2024.

\bibitem{DBLP:conf/icassp/YangHCKT23}
Karren~D. Yang, Ting{-}Yao Hu, Jen{-}Hao~Rick Chang, Hema~Swetha Koppula, and Oncel Tuzel,
\newblock ``Text is all you need: Personalizing {ASR} models using controllable speech synthesis,''
\newblock in {\em Proc. of ICASSP}, 2023.

\bibitem{DBLP:conf/icassp/0028HWGLG20}
Yan Huang, Lei He, Wenning Wei, William Gale, Jinyu Li, and Yifan Gong,
\newblock ``Using personalized speech synthesis and neural language generator for rapid speaker adaptation,''
\newblock in {\em Proc. of ICASSP}, 2020.

\bibitem{DBLP:conf/interspeech/0028LHWGG20}
Yan Huang, Jinyu Li, Lei He, Wenning Wei, William Gale, and Yifan Gong,
\newblock ``Rapid {RNN-T} adaptation using personalized speech synthesis and neural language generator,''
\newblock in {\em Proc. of Interspeech}, Helen Meng, Bo~Xu, and Thomas~Fang Zheng, Eds., 2020.

\bibitem{yang2024enhancing}
Guanrou Yang, Fan Yu, Ziyang Ma, Zhihao Du, Zhifu Gao, Shiliang Zhang, and Xie Chen,
\newblock ``Enhancing low-resource asr through versatile tts: Bridging the data gap,''
\newblock {\em arXiv preprint arXiv:2410.16726}, 2024.

\bibitem{DBLP:conf/icassp/SoleymanpourJSB22}
Mohammad Soleymanpour, Michael~T. Johnson, Rahim Soleymanpour, and Jeffrey Berry,
\newblock ``Synthesizing dysarthric speech using multi-speaker tts for dysarthric speech recognition,''
\newblock in {\em Proc. of ICASSP}, 2022.

\bibitem{gpt2023gpt}
OpenAI,
\newblock ``{GPT-4} technical report,''
\newblock {\em arXiv preprint arXiv:2303.08774}, 2023.

\bibitem{zeng2023glm130b}
Aohan Zeng, Xiao Liu, Zhengxiao Du, Zihan Wang, Hanyu Lai, Ming Ding, Zhuoyi Yang, Yifan Xu, Wendi Zheng, Xiao Xia, Weng~Lam Tam, Zixuan Ma, Yufei Xue, Jidong Zhai, Wenguang Chen, Peng Zhang, Yuxiao Dong, and Jie Tang,
\newblock ``Glm-130b: An open bilingual pre-trained model,''
\newblock {\em Proc. of ICLR}, 2023.

\bibitem{touvron2023llama}
Hugo Touvron, Thibaut Lavril, Gautier Izacard, Xavier Martinet, Marie-Anne Lachaux, Timoth{\'e}e Lacroix, Baptiste Rozi{\`e}re, Naman Goyal, Eric Hambro, Faisal Azhar, et~al.,
\newblock ``{LLaMA}: Open and efficient foundation language models,''
\newblock {\em arXiv preprint arXiv:2302.13971}, 2023.

\bibitem{ma2023generative}
Rao Ma, Mengjie Qian, Potsawee Manakul, Mark Gales, and Kate Knill,
\newblock ``Can generative large language models perform asr error correction?,''
\newblock {\em arXiv preprint arXiv:2307.04172}, 2023.

\bibitem{Shu_2024}
Lei Shu, Liangchen Luo, Jayakumar Hoskere, Yun Zhu, Yinxiao Liu, Simon Tong, Jindong Chen, and Lei Meng,
\newblock ``Rewritelm: An instruction-tuned large language model for text rewriting,''
\newblock {\em Proc. AAAI}, vol. 38, no. 17, pp. 18970–18980, Mar. 2024.

\bibitem{Fan_2023}
Yaxin Fan, Feng Jiang, Peifeng Li, and Haizhou Li,
\newblock {\em GrammarGPT: Exploring Open-Source LLMs for Native Chinese Grammatical Error Correction with Supervised Fine-Tuning}, p. 69–80,
\newblock Springer Nature Switzerland, 2023.

\bibitem{alharbi2024leveraging}
Sadeen Alharbi, Reem BinMuqbil, Ahmed Ali, Raghad AlOraini, Saiful Bari, Areeb Alowisheq, and Yaser Alonaizan,
\newblock ``Leveraging llm for augmenting textual data in code-switching asr: Arabic as an example,''
\newblock in {\em Proc. SynData4GenAI}, 2024.

\bibitem{DBLP:journals/corr/abs-2408-09215}
Samuele Cornell, Jordan Darefsky, Zhiyao Duan, and Shinji Watanabe,
\newblock ``Generating data with text-to-speech and large-language models for conversational speech recognition,''
\newblock {\em CoRR}, vol. abs/2408.09215, 2024.

\bibitem{DBLP:conf/icassp/SuHKVCYMT24}
Hsuan Su, Ting{-}Yao Hu, Hema~Swetha Koppula, Raviteja Vemulapalli, Jen{-}Hao~Rick Chang, Karren~D. Yang, Gautam~Varma Mantena, and Oncel Tuzel,
\newblock ``Corpus synthesis for zero-shot {ASR} domain adaptation using large language models,''
\newblock in {\em Proc. of ICASSP}, 2024.

\bibitem{panayotov2015librispeech}
Vassil Panayotov, Guoguo Chen, Daniel Povey, and Sanjeev Khudanpur,
\newblock ``Librispeech: an asr corpus based on public domain audio books,''
\newblock in {\em Proc. of ICASSP}, {South Brisbane, Queensland, Australia}, {Apr.} 2015, pp. 5206--5210.

\bibitem{vaswani2017attention}
Ashish Vaswani, Noam Shazeer, Niki Parmar, Jakob Uszkoreit, Llion Jones, Aidan~N Gomez, {\L}ukasz Kaiser, and Illia Polosukhin,
\newblock ``Attention is all you need,''
\newblock in {\em Proc. NeurIPS}, {Long Beach, California, USA}, {Dec.} 2017.

\bibitem{gulati2020conformer}
A.~Gulati, J.~Qin, C.~Chiu, N.~Parmar, Y.~Zhang, J.~Yu, W.~Han, S.~Wang, Z.~Zhang, Y.~Wu, et~al.,
\newblock ``Conformer: Convolution-augmented transformer for speech recognition,''
\newblock {\em Proc. of Interspeech}, 2020.

\bibitem{watanabe2018espnet}
Shinji Watanabe, Takaaki Hori, Shigeki Karita, Tomoki Hayashi, Jiro Nishitoba, Yuya Unno, Nelson {Enrique Yalta Soplin}, Jahn Heymann, Matthew Wiesner, Nanxin Chen, Adithya Renduchintala, and Tsubasa Ochiai,
\newblock ``{ESPnet}: End-to-end speech processing toolkit,''
\newblock in {\em Proc. of Interspeech}, 2018, pp. 2207--2211.

\bibitem{park2019specaugment}
Daniel~S Park, William Chan, Yu~Zhang, Chung-Cheng Chiu, Barret Zoph, Ekin~D Cubuk, and Quoc~V Le,
\newblock ``Specaugment: A simple data augmentation method for automatic speech recognition,''
\newblock in {\em Proc. of Interspeech}, {Graz, Austria}, {Sep.} 2019, pp. 2613--2617.

\bibitem{watanabe2017hybrid}
Shinji Watanabe, Takaaki Hori, Suyoun Kim, John~R Hershey, and Tomoki Hayashi,
\newblock ``Hybrid {CTC}/attention architecture for end-to-end speech recognition,''
\newblock {\em IEEE Journal of Selected Topics in Signal Processing}, vol. 11, no. 8, pp. 1240--1253, 2017.

\bibitem{mosnet}
Chen-Chou Lo, Szu-Wei Fu, Wen-Chin Huang, Xin Wang, Junichi Yamagishi, Yu~Tsao, and Hsin-Min Wang,
\newblock ``Mosnet: Deep learning based objective assessment for voice conversion,''
\newblock in {\em Proc. of Interspeech}, 2019.

\bibitem{chen2021wavlm}
S.~Chen, C.~Wang, Z.~Chen, Y.~Wu, S.~Liu, Z.~Chen, J.~Li, N.~Kanda, T.~Yoshioka, X.~Xiao, et~al.,
\newblock ``{WavLM}: Large-scale self-supervised pre-training for full stack speech processing,''
\newblock {\em IEEE Journal of Selected Topics in Signal Processing}, 2022.

\bibitem{feng2021quantifying}
Siyuan Feng, Olya Kudina, Bence~Mark Halpern, and Odette Scharenborg,
\newblock ``Quantifying bias in automatic speech recognition,''
\newblock {\em arXiv preprint arXiv:2103.15122}, 2021.

\bibitem{radford2022robust}
Alec Radford, Jong~Wook Kim, Tao Xu, Greg Brockman, Christine McLeavey, and Ilya Sutskever,
\newblock ``Robust speech recognition via large-scale weak supervision,''
\newblock {\em arXiv preprint arXiv:2212.04356}, 2022.

\end{thebibliography}

\end{document}